\documentclass[sigconf]{acmart}

\usepackage{booktabs} 
\usepackage{algorithmic}
\usepackage{graphicx}

\usepackage{subfigure}
\usepackage{multirow}
\usepackage{pifont}
\newcommand{\cmark}{\ding{51}}%
\newcommand{\xmark}{\ding{55}}%
\usepackage{enumitem}
\usepackage{scalerel}
\usepackage{hyperref}

\setcopyright{acmcopyright}

\copyrightyear{2022}
\acmYear{2022}
\setcopyright{acmcopyright}\acmConference[SAC '22]{The 37th ACM/SIGAPP Symposium on Applied Computing}{April 25--29, 2022}{Virtual Event, }
\acmBooktitle{The 37th ACM/SIGAPP Symposium on Applied Computing (SAC '22), April 25--29, 2022, Virtual Event, }
\acmPrice{15.00}
\acmDOI{10.1145/3477314.3508380}
\acmISBN{978-1-4503-8713-2/22/04}

\acmArticle{4}

\begin{document}
\title{Improving Response Time of Home IoT Services in Federated Learning}
  
\renewcommand{\shorttitle}{SIG Proceedings Paper in LaTeX Format}

\author{Dongjun Hwang}
\authornotemark[1]
\orcid{0000-0001-8891-7654}
\affiliation{%
  \institution{Chungnam National University}
  \streetaddress{99, Daehak-ro 76beon-gil, Yuseong-gu}
  \city{Daejeon}
  \postcode{34183}
  \country{Republic of Korea}
}
\email{enter031@cnu.ac.kr}

\author{Hyunsu Mun}
\affiliation{%
  \institution{Chungnam National University}
  \streetaddress{99, Daehak-ro 76beon-gil, Yuseong-gu}
  \city{Daejeon}
  \postcode{34183}
  \country{Republic of Korea}
}
\email{munhyunsu@cnu.ac.kr}

\author{Youngseok Lee}
\affiliation{%
  \institution{Chungnam National University}
  \streetaddress{99, Daehak-ro 76beon-gil, Yuseong-gu}
  \city{Daejeon}
  \postcode{34183}
  \country{Republic of Korea}
}
\email{lee@cnu.ac.kr}

\renewcommand{\shortauthors}{D. Hwang et al.}

\begin{abstract}
For intelligent home IoT services with sensors and machine learning, we need to upload IoT data to the cloud server which cannot share private data for training.
A recent machine learning approach, called federated learning, keeps user data on the device in the distributed computing environment.
Though federated learning is useful for protecting privacy, it experiences poor performance in terms of the end-to-end response time in home IoT services, because IoT devices are usually controlled by remote servers in the cloud.
In addition, it is difficult to achieve the high accuracy of federated learning models due to insufficient data problems and model inversion attacks.
In this paper, we propose a local IoT control method for a federated learning home service that recognizes the user behavior in the home network quickly and accurately.
We present a federated learning client with transfer learning and differential privacy to solve data scarcity and data model inversion attack problems.
From experiments, we show that the local control of home IoT devices for user authentication and control message transmission by the federated learning clients improves the response time to less than 1 second.
Moreover, we demonstrate that federated learning with transfer learning achieves 97\% of accuracy under 9,000 samples, which is only 2\% of the difference from centralized learning.
\end{abstract}

%
%



\maketitle

\section{Introduction}
\label{sec:introduction}
Internet of Things (IoT) equipped with sensors and machine learning has been explosively popular\footnote{https://www.statista.com/statistics/485136/global-internet-of-things-market-size/}.
Home IoT devices such as built-in sensors, cameras, light bulbs, speakers, door locks, or window chains are managed by a smartphone for automation services.
Intelligent IoT services increase the efficiency and the convenience to users.
For example, a smart bulb like Philips Hue can change light colors \cite{rebecca2021lightblub}.
In addition, a smart speaker such as Google Nest Hub is connected to IoT devices controlled by users through voice commands \cite{molly2021nesthub}.

Home IoT services are often vulnerable to privacy problems because they can be accessed from a remote server in the cloud and their data in the cloud can be exposed to the external attacks.
As home IoT data contains personal information, it is difficult to share the private data publicly.
Today, many countries have laws or regulations to protect privacy.
General Data Protection Regulation (GDPR) is issued by the European Union for data privacy and security \cite{voigt2017eu}.
In particular, as most IoT services depend on the centralized cloud, information leakage might be possible.
It is an important challenge to address the personal data protection in home IoT services.

A recent machine learning approach, called federated learning (FL), protects user data by keeping them on the device in a distributed computing environment.
In the federated learning model, each client performs the local learning job on the device, and then it uploads only the parameters of the local model to the FL server.
A FL server aggregates all parameters to compile the new global model.
Federated learning protects privacy because only the parameters of the model are shared.
Google demonstrate Gboard application in federated learning to predict next words typed by a user \cite{hard2018federated}.
As words are stored on the device, there is no risk of data leakage to the outside.

Yet, home IoT services with federated learning meet challenges in maximizing the user satisfaction: the response time of IoT devices controlled by a cloud server is slow; the accuracy of the model is not high because of insufficient data; threats to privacy are possible from a model inversion attack.
The response time is one of the important factors to user experience.
However, the response time of IoT services becomes slow when devices are controlled by the cloud server.
In addition, insufficient data of the home network is the cause of lowering the accuracy of the training model.
Model inversion attacks can extract training data from parameters between FL clients and a server.

In this paper, we propose a local IoT control method for federated learning home IoT service.
We minimize the response time by employing the local control of IoT devices method.
We also improve the accuracy of FL under insufficient data by applying transfer learning and mitigate private information leakage through model inversion attacks on the FL client with differential privacy.

From experiments, we show that local control of home IoT devices reduce the end-to-end response time by up to 18\% when compared to centralized learning (CL).
Our method provides the fast intelligent IoT service within 1 second.
Furthermore, federated learning with transfer learning achieves the accuracy of 97\% under about 9,000 samples, which is only 2\% different from centralized learning.
The accuracy of federated learning with differential privacy is 93\%, which is 4\% difference compared to the case without differential privacy.

\section{Related Work}
\begin{table}
  \caption{Comparison between related work and our proposal.}
  \label{tab:summary_reference_works}
  \centering
  \resizebox{0.48\textwidth}{!}{%
  \begin{tabular}{lcccc}
    \toprule
    & \multirow[t]{2}{*}{IoT Control} & \multirow[t]{2}{*}{Federated} &
    \multirow[t]{2}{*}{Transfer} &
    \multirow[t]{2}{*}{Differential} \\
    & \multirow[t]{2}{*}{Service} & \multirow[t]{2}{*}{Learning} &
    \multirow[t]{2}{*}{Learning} &
    \multirow[t]{2}{*}{Privacy} \\
    \midrule
    Aivodji \textit{et al.} \cite{aivodji2019iotfla} & Remote & \cmark & \xmark & \xmark \\
    Wu \textit{et al.} \cite{wu2020personalized} & Remote & \cmark & \xmark & \xmark \\
    Cao \textit{et al.} \cite{cao2020ifed} & Remote & \cmark & \xmark & \cmark \\
    Rodriguez \textit{et al.} \cite{rodriguez2020federated} & Remote & \cmark & \xmark & \cmark \\
    Our approach & Local & \cmark & \cmark & \cmark \\
    \bottomrule
\end{tabular}}
\end{table}

\textbf{Response time of IoT service:} 
In \cite{9424693}, the authors used edge computing environments with SDN networks to reduce the response time of IoT applications.
\cite{bouloukakis2021performance} presents different types of queuing models for QoS settings of IoT device interactions, showing that they have a significant impact on delivery success rate and response time.
\cite{8468984} proposes a service cache policy by utilizing the combinability of services to improve the performance of the service providing system.
The author states that the average response time of IoT services can be improved as a result of conducting a series of experiments to evaluate the performance of the approach.

In our previous study, we measured and analyzed response times for IoT devices with and without cloud environments \cite{lee2020comparing}.
In \cite{kumer2021smart}, the author present context-aware IoT services in remote control.
They use the trigger-action third-party IoT management service, IFTTT. 
However, the use of IFTTT cloud servers when controlling IoT devices often results in long response time.

\textbf{IoT service in federated learning:} \cite{aivodji2019iotfla} and \cite{wu2020personalized} propose a personalized federated learning framework for protecting user privacy in the existing IoT service environment.
Rodríguez-Barroso \textit{et al.} \cite{rodriguez2020federated} and Cao \textit{et al.} \cite{cao2020ifed} applied differential privacy to the existing federated learning framework for privacy protection.

Table \ref{tab:summary_reference_works} compares related work with our proposed method.
We examine the bottleneck of the slow response time and improve the latency of federated learning IoT control.
In addition, we support transfer learning and differential privacy together to improve the accuracy of FL.
Previous studies applied federated learning to IoT, but they did not consider the response time.

\section{Accelerating Federated Learning Home IoT Service}
\subsection{Home IoT Service in Federated Learning}
\begin{figure*}[htb]
    \centering
    \includegraphics[width=0.96\textwidth]{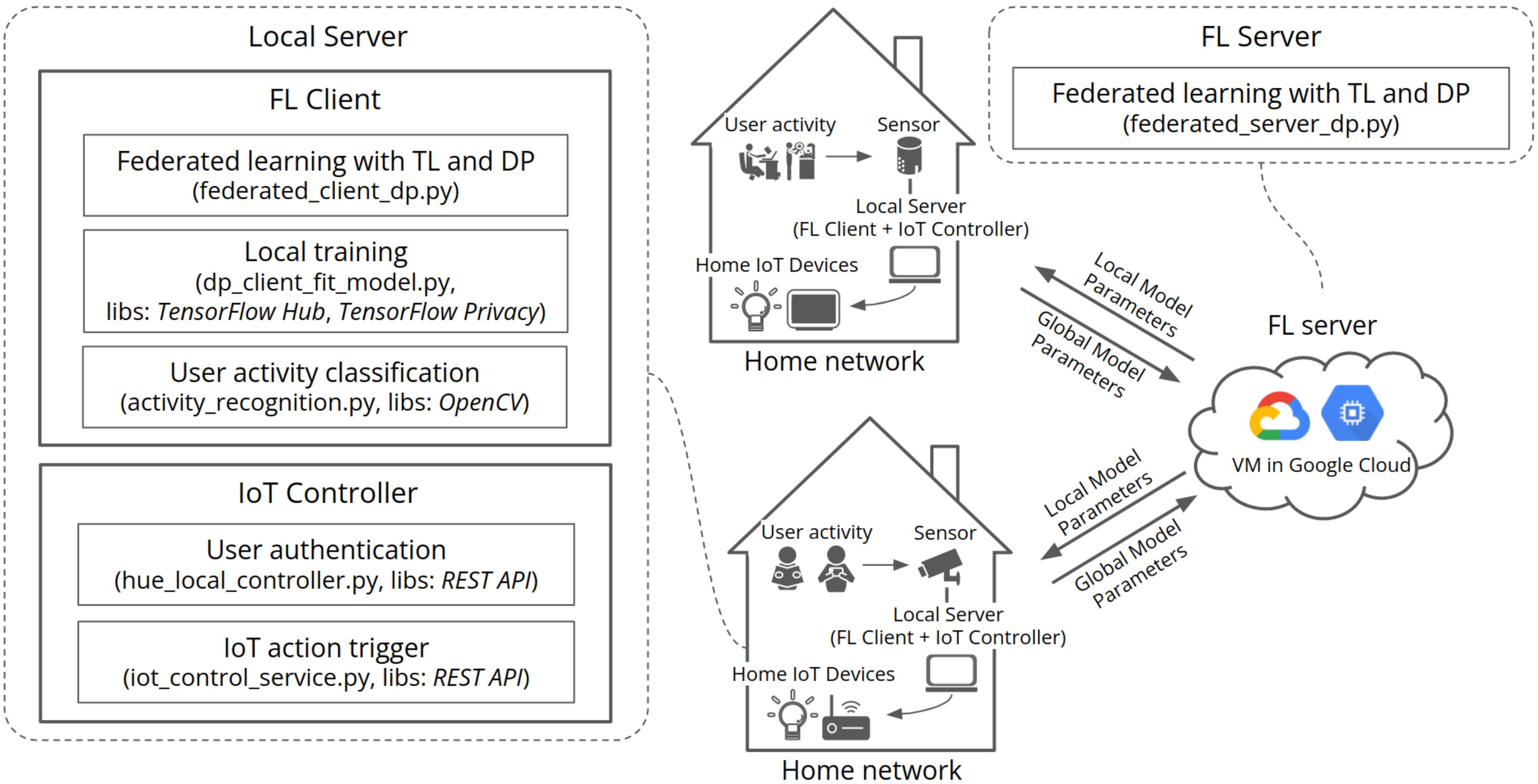}
    \caption{The overview of home IoT service with local control in federated learning.}
    \label{fig:overview}
\end{figure*}

\begin{table}
  \caption{IoT service scenarios according to user activity.}
  \label{tab:context_service_device}
  \centering
  \resizebox{0.48\textwidth}{!}{%
  \begin{tabular}{ll}
    \toprule
    User activity & IoT service scenario \\
    \midrule
    Reading & Turn on smart light \\
    \multirow[t]{2}{*}{Drinking water} & \multirow[t]{2}{*}{Record user water intake in the local server database} \\ 
    & \multirow[t]{2}{*}{and notify with smart speaker} \\
    Using laptop & Block harmful URL at WiFi router \\
    Using mobile phone & Manage traffic at WiFi router \\
    Washing dishes & Play YouTube with smart speaker \\
  \bottomrule
\end{tabular}}
\end{table}

Figure \ref{fig:overview} is the overview of the home IoT service in federated learning.
First, the FL client performs the local learning job using sensor data to detect user activities.
For local training, the FL client communicates with the FL server.
We combine the federated learning model with transfer learning to compensate for insufficient data.
Additionally, we have enhanced privacy protection from model inversion attacks by applying differential privacy to our training model.

The local server (FL client + IoT controller) controls the home IoT device suited for the scenario corresponding to the classified activity.
We apply transfer learning (TL) and differential privacy (DP) to the federated learning model in the local training process in FL client.
As the model trained through federated learning resides on the local server, the FL client does not need to communicate with the server for the classification job.
The IoT controller on the local server manages the IoT device according to the classified user activity.
The IoT controller authenticates users and sends control messages directly to the IoT device for home services.
In Table \ref{tab:context_service_device}, we summarize user activities and the corresponding IoT services.

\subsection{IoT Device Control: Local vs. Remote}
Home IoT services typically require servers to perform complex tasks such as connecting IoT devices and generating control commands through machine learning models.
In centralized learning, a cloud server trains a machine learning model for home IoT services, and the inference process is also performed on the server.
IoT devices are remotely controlled through centralized learning.
On the other hand, federated learning runs machine learning models on a local device, minimizing communication with remote servers.
Therefore, local control of home IoT devices through federated learning reduces the communication process with the server to the minimum, enabling fast service within a short time.

\begin{figure}[htb]
    \centering
    \includegraphics[width=0.48\textwidth]{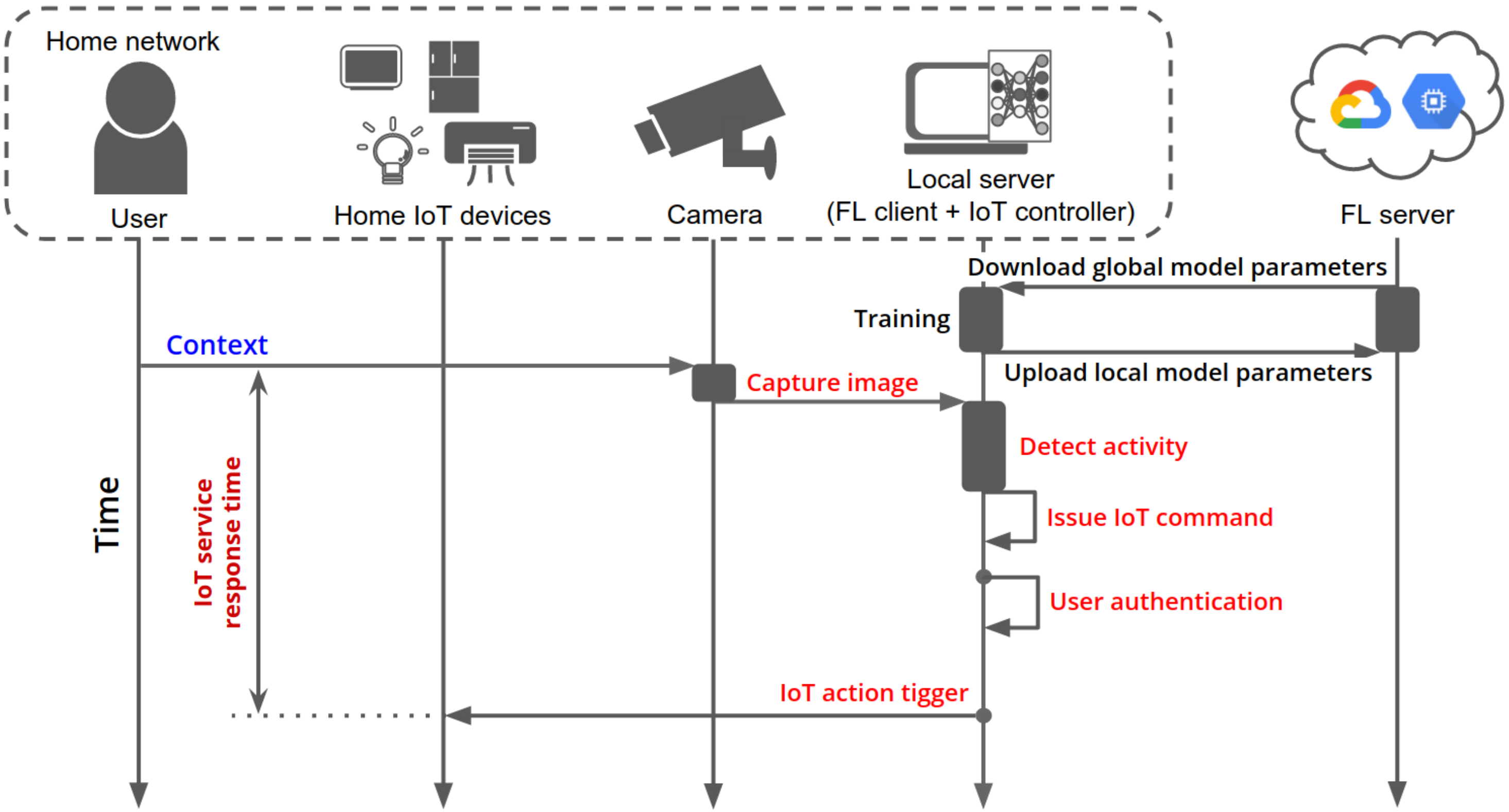}
    \caption{Local control of home IoT service with the federated learning model.}
    \label{fig:fl_control_flow}
\end{figure}

Figure \ref{fig:fl_control_flow} shows how local control is combined with federated learning.
We assume a home network consisting of sensors, a local server (FL client and IoT controller), and IoT devices.
The FL client detects user activities through federated learning.
The IoT controller authenticates the user allowed to control the device, and controls the device based on the classified activity.
Our home IoT service, combined with federated learning and local control, can quickly improve the response time by performing all processes locally from data analysis to user authentication and control.

\begin{figure}[htb]
    \centering
    \includegraphics[width=0.48\textwidth]{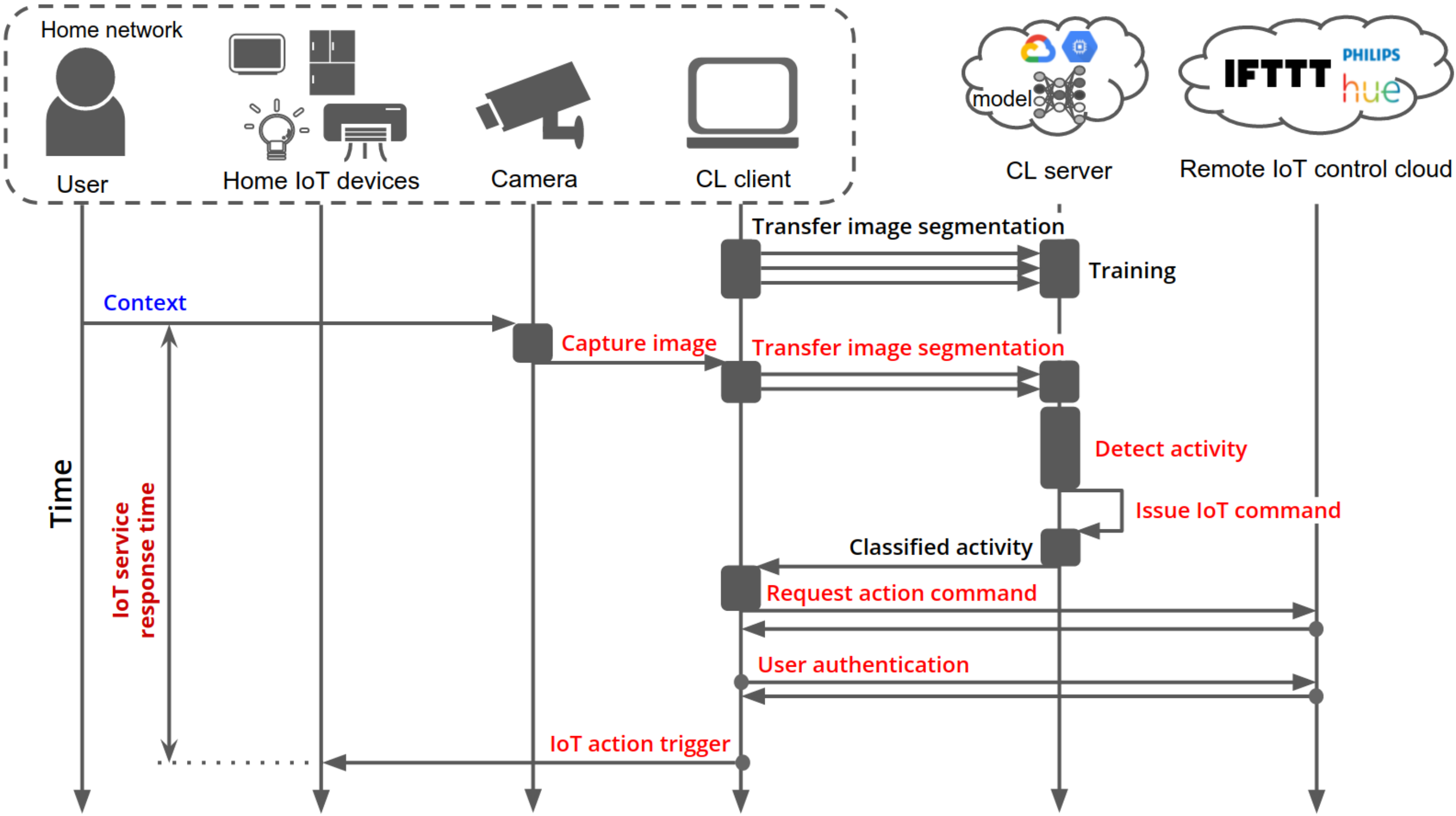}
    \caption{Remote control (IFTTT) of home IoT service with the centralized learning model.}
    \label{fig:cl_control_flow}
\end{figure}

Figure \ref{fig:cl_control_flow} shows an example of remote IoT control via a cloud server.
The CL client sends data to the CL server to classify the user activity captured by a camera.
The third-party IoT service such as IFTTT\footnote{https://ifttt.com/} provides the remote IoT authentication and control service.
Centralized learning puts the training model in the cloud to analyze the received images.
In addition to the increased communication latency of the CL server, authentication via cloud IFTTT makes the response time slow.

\subsection{Federated Learning with Transfer Learning and Differential Privacy}
\begin{figure}[htb]
    \centering
    \includegraphics[width=0.48\textwidth]{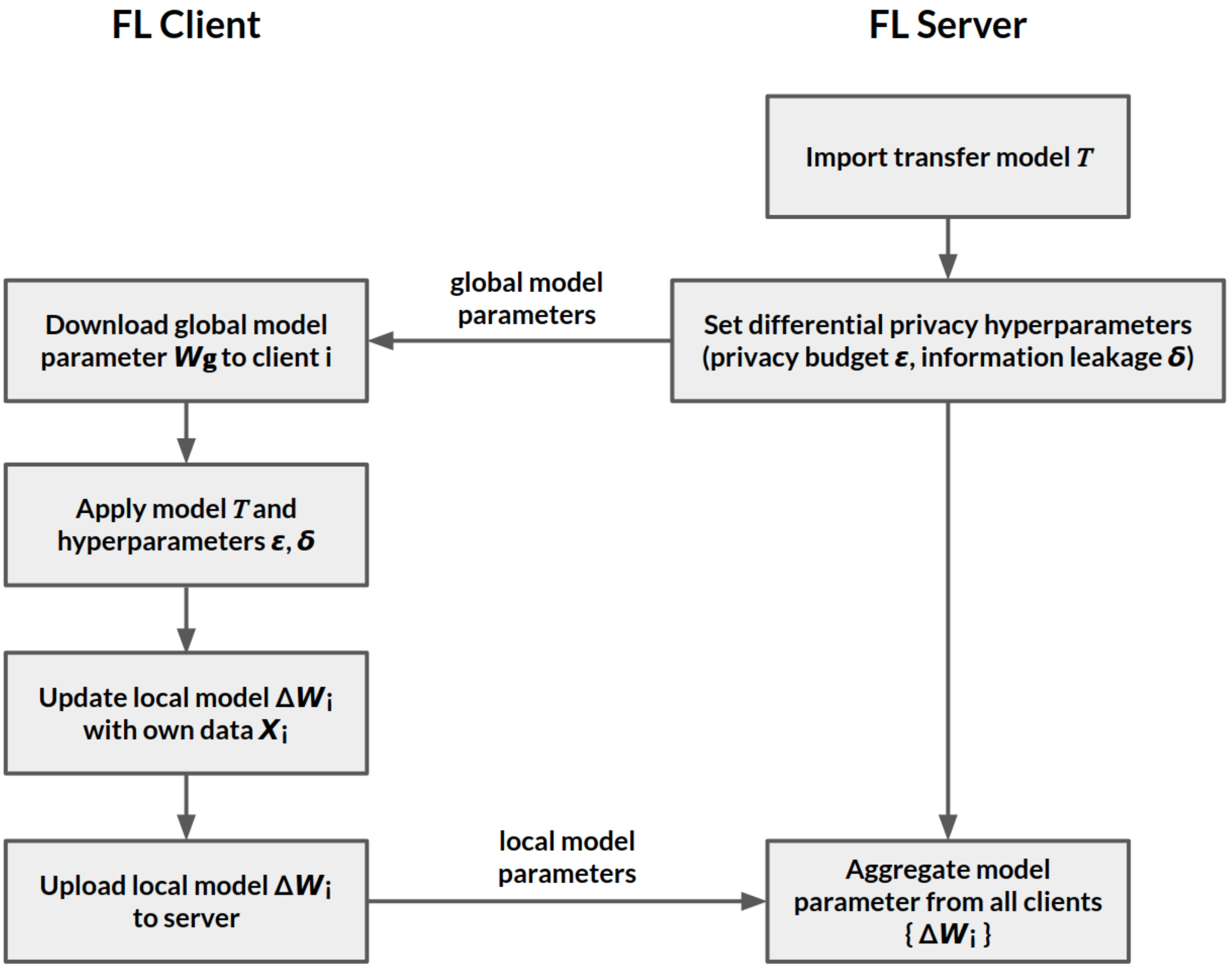}
    \caption{The flowchart of federated learning with TL and DP.}
    \label{fig:fl_train}
\end{figure}

We combine federated learning with transfer learning and differential privacy to improve model performance against insufficient data and enhance privacy protection against model inversion attacks.
Federated learning combined with TL and DP is shown in Fig. \ref{fig:fl_train}.
Before starting training, a FL server in the cloud imports the transfer model, $T$.
The FL server sets the initial value of $\epsilon$ and $\delta$, which are hyperparameters for differential privacy.
FL client $i$ applies the global model parameter $W_g$ downloaded from a FL server and hyperparameter to model, $T$.
In the next step, the FL client $i$ updates the parameters of the local model $\Delta W_i$ based on the data $X_i$ and the model parameter $W_i$.
Each FL client then uploads the updated local model parameter $\Delta W_i$ to the FL server.
Finally, the FL server aggregates the parameters ${ \Delta W_i }$ for all clients.
The training process builds the model iteratively in each round.
In federated learning, the transfer model, $T$ learns the characteristics of training data in advance, and it solves the insufficient data for each client.
The model inversion attack can estimate the training data $X_i$ using the parameter $\Delta W_i$ of the model.
We add noise to the parameters via the differential privacy parameters $\epsilon$, $\delta$.

\section{Experiments} \label{experiments}
\subsection{Experiment Environment}
\begin{figure}
    \centering
    \includegraphics[width=0.48\textwidth]{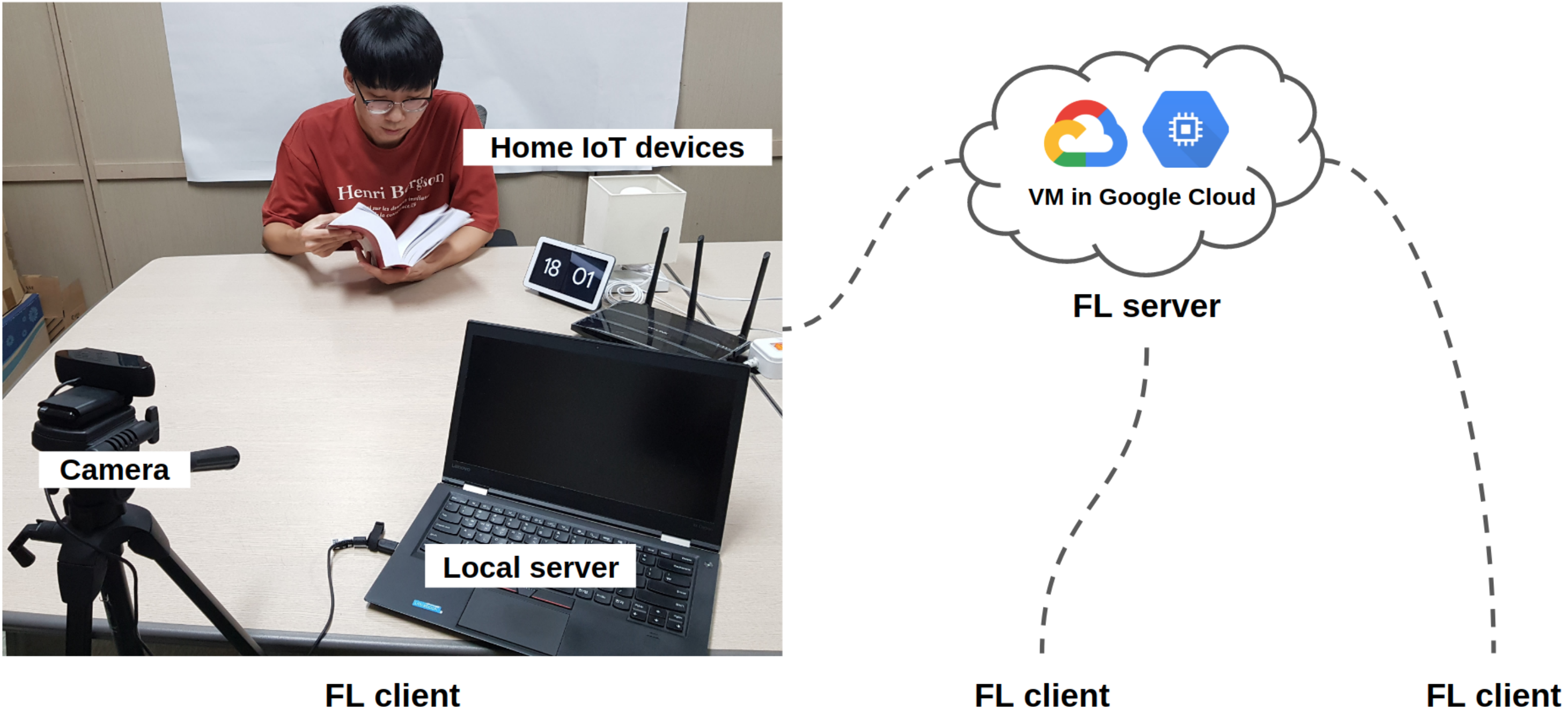}
    \caption{The testbed of local IoT control for federated learning home service.}
    \label{fig:testbed}
\end{figure}

Figure \ref{fig:testbed} is the experiment environment of home IoT service with federated learning.
We have implemented a home IoT service in federated learning with Tensorflow\footnote{https://www.tensorflow.org/} and OpenCV\footnote{https://opencv.org/}.
We configure a FL server with VM in Google Cloud.
FL clients and a FL server communicate with WebSocket\footnote{https://websockets.readthedocs.io/en/stable/}.
We connect the controller of IoT devices such as Philips Hue, TP-Link, and Google Nest Hub to a local server with a camera.
The source code is available on GitHub\footnote{\url{https://github.com/HwangDongJun/Federated\_Learning\_using\_Websockets}}.
We summarize the components in our experiment as follows.

\begin{itemize}
    \item Camera: webcam (Logitech C920).
    \item Local Server (FL client + IoT controller): laptop (Lenovo ThinkPad X1) in Ubuntu 20.04 LTS.
    \item IoT devices: smart light (Philips Hue), WiFi router (TP-Link), and a smart speaker (Google Nest Hub).
\end{itemize}

\begin{table}
  \caption{Learning model parameters by round.}
  \label{tab:model_parameters}
  \centering
  \resizebox{0.48\textwidth}{!}{%
  \begin{tabular}{lccc}
    \toprule
    \multirow[t]{2}{*}{Parameter} & \multirow[t]{2}{*}{1$_{st}$ round} & \multirow[t]{2}{*}{1$_{st}$ round} & \multirow[t]{2}{*}{n$_{th}$ round (n$\geq$2)} \\
    & \multirow[t]{2}{*}{(Freeze the pre-trained layer)} & \multirow[t]{2}{*}{(Fine-tune the model)} & \\
    \midrule
    Epochs & 10 & 30 & 10 \\
    Learning rate & $10^{-3}$ & $10^{-4}$ & $10^{-4}$ \\
    Batch size & 32 & 32 & 32 \\
  \bottomrule
\end{tabular}}
\end{table}

\textbf{Building a model:} For experiments, we use models of MobilenetV2 \cite{howard2018inverted}, and EfficientnetB0 \cite{tan2019efficientnet}.
Both models are initially trained with an input image of size 224 $\times$ 224 $\times$ 3.
We describe the parameters required for model training in Table \ref{tab:model_parameters}.
To initially obtain a learning baseline, we train transfer learning model.
In the 1$_{st}$ round, the conv2D and dense layers are not updated during training, only the weights of the softmax layer that have been changed to match the new class are updated.
In other words, all layers are set to be frozen except the last softmax layer.
The initial learning rate is set to $10^{-3}$ and the model is trained for 10 epochs.
After training the last softmax layer, we fine-tune the training model.
We train the fine-tuned model by changing the epoch to 30 and the learning rate to $10^{-4}$.
After the 1$_{st}$ round, the model trains for 10 epochs.
We limit the epochs to avoid overfitting because the overall amount of data is small and we reuse the model trained in the previous round.

\begin{table}
  \caption{Dataset of five activity images.}
  \label{tab:using learning image data}
  \centering
  \resizebox{0.48\textwidth}{!}{%
  \begin{tabular}{lcccccc}
    \toprule
    User & Reading & Drinking Water & Using Laptop & Using Mobile Phone & Washing Dishes & Total\\
    \midrule
    A & 518 & 517 & 504 & 564 & 502 & 2,605 \\
    B & 433 & 371 & 529 & 288 & 422 & 2,043 \\
    C & 478 & 372 & 527 & 433 & 462 & 2,272 \\
  Total & 1,429 & 1,260 & 1,560 & 1,285 & 1,386 & 6,920 \\ 
  \bottomrule
\end{tabular}}
\end{table}

\textbf{Datasets:} We collect 8,948 image frames through the camera for the five activity categories discussed earlier in Table \ref{tab:context_service_device}.
For the accuracy test, after recording a video file for three seconds in 10 frames per second, we label the corresponding action for each image.
Data is divided into training and test dataset as shown in Table \ref{tab:using learning image data}.
We set up three participants for the experiment and collect five activity images.
We have 6,920 frames of training data and 2,028 frames of test data.

\subsection{Response Time}
\begin{figure}[htb]
    \centering
    \subfigure[Local control with FL]{
        \includegraphics[width=0.98\columnwidth]{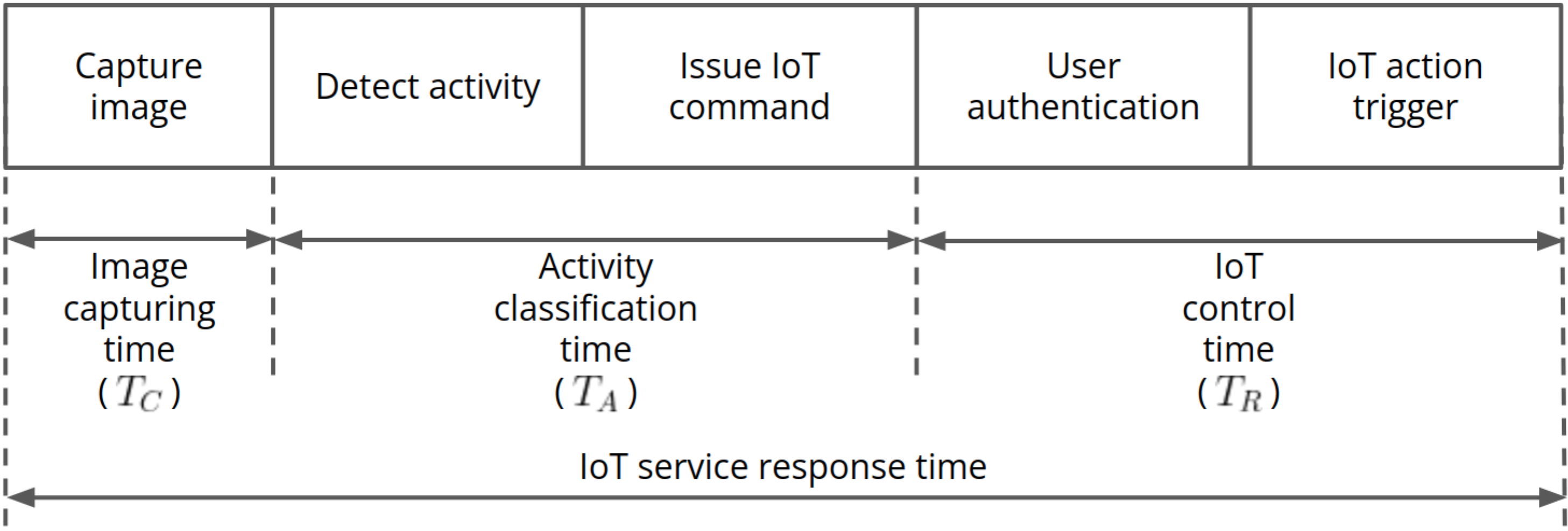}
    }
    \centering
    \subfigure[Remote control with CL]{
        \includegraphics[width=0.98\columnwidth]{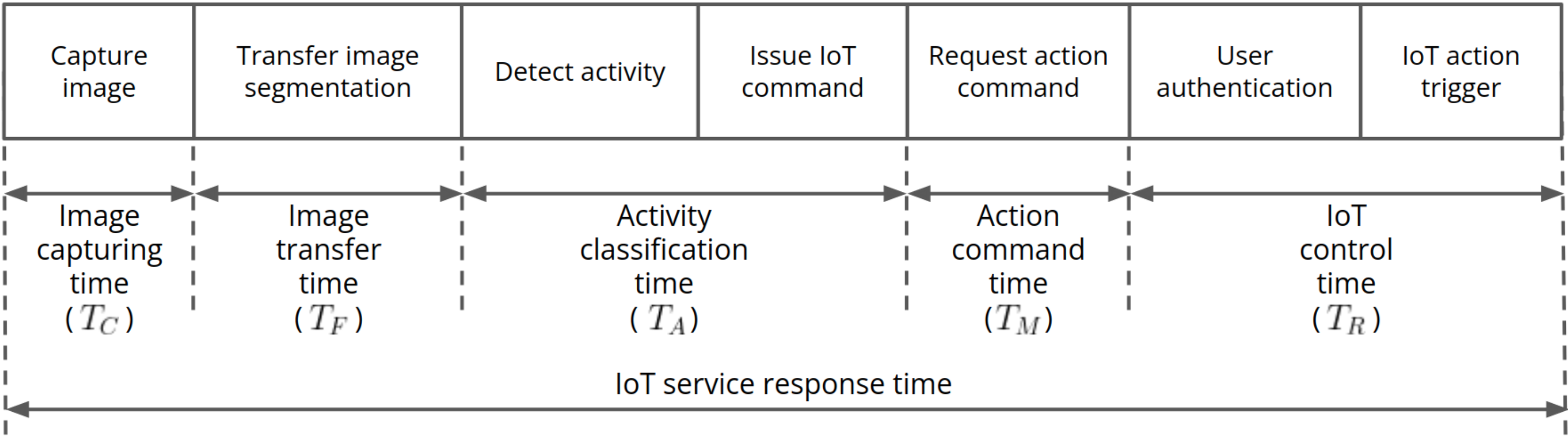}
    }
    \caption{Breakdown of IoT service response time.}
    \label{fig:fl_cl_response_time}
\end{figure}

We investigate the end-to-end response time of home IoT services consisting of local control and federated learning steps.
We compare local control with FL and remote control with CL as an IoT services.
The response time is the time between capturing image and control an IoT device.
Figure \ref{fig:fl_cl_response_time} shows the end-to-end IoT service response time consisting of capturing, transmitting images, and detecting user activities, user authentication, and issuing IoT action trigger commands.


\begin{figure}
    \centering
    \includegraphics[width=0.45\textwidth]{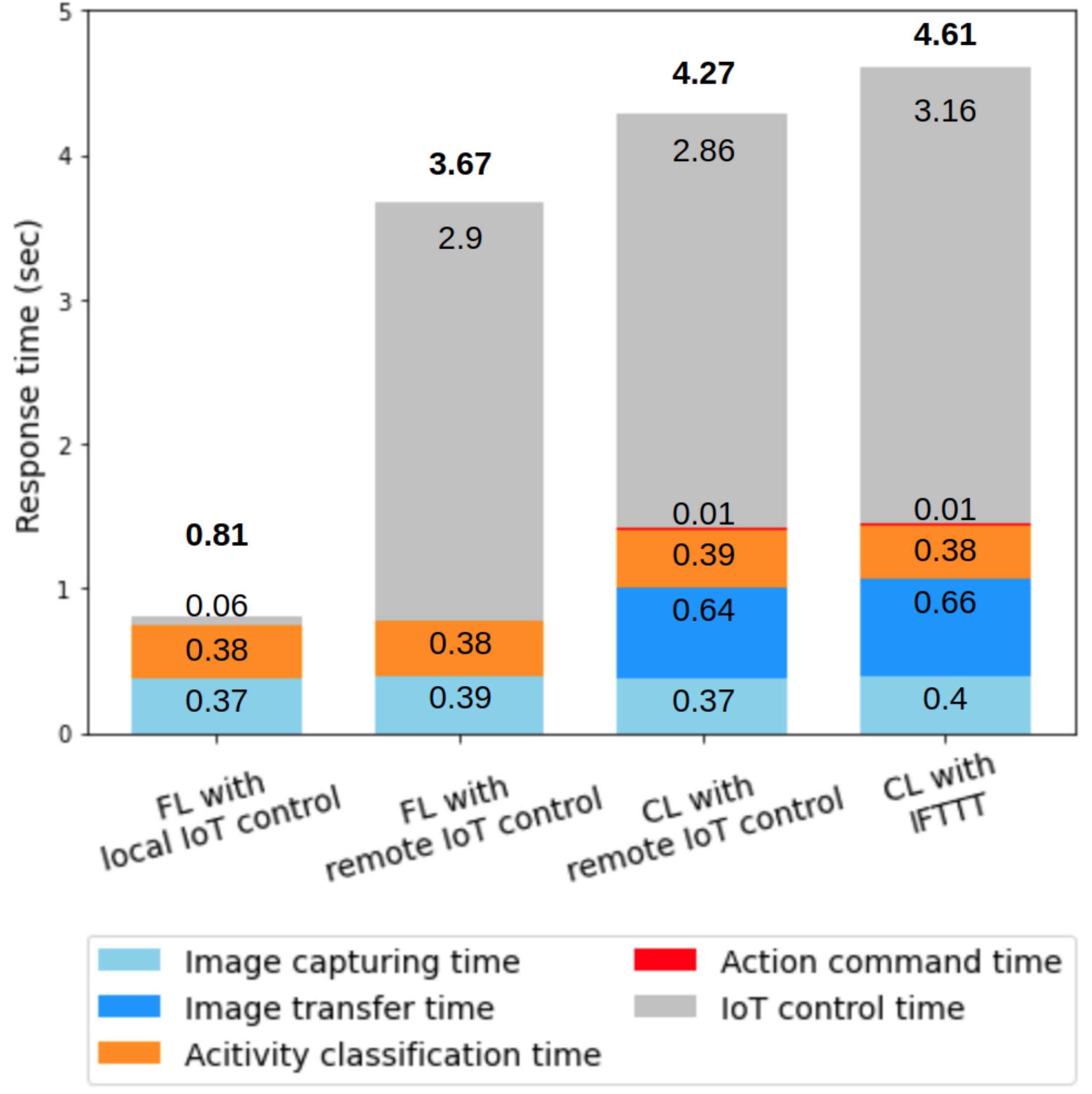}
    \caption{The response time of detecting a user activity (reading) and turning a smart light on: FL vs. CL under local and remote controls.}
    \label{fig:cl_vs_lo_time}
\end{figure}

Figure \ref{fig:cl_vs_lo_time} shows the IoT service response time from $T_C$ to $T_R$.
We compare FL and CL under local, remote, or IFTTT for smart light control.
The service response time with FL and local IoT control is only 0.81 seconds.
However, the response time increases to 3.67 seconds with FL in remote IoT control, and 4.27 seconds with CL in remote IoT control.
Remote IoT control from CL using IFTTT has the response time of 4.61 seconds.
In CL with remote IoT control, it takes 0.64 seconds to transfer image to the server, and 2.86 seconds for the cloud server to trigger an action to the IoT device, which is the bottleneck of the IoT control.
In the case of IFTTT, it takes a long time (3.16 seconds) for the IoT control because the authentication and IoT control are performed through the IFTTT server and the remote IoT server.

\begin{figure}
    \centering
    \includegraphics[width=0.45\textwidth]{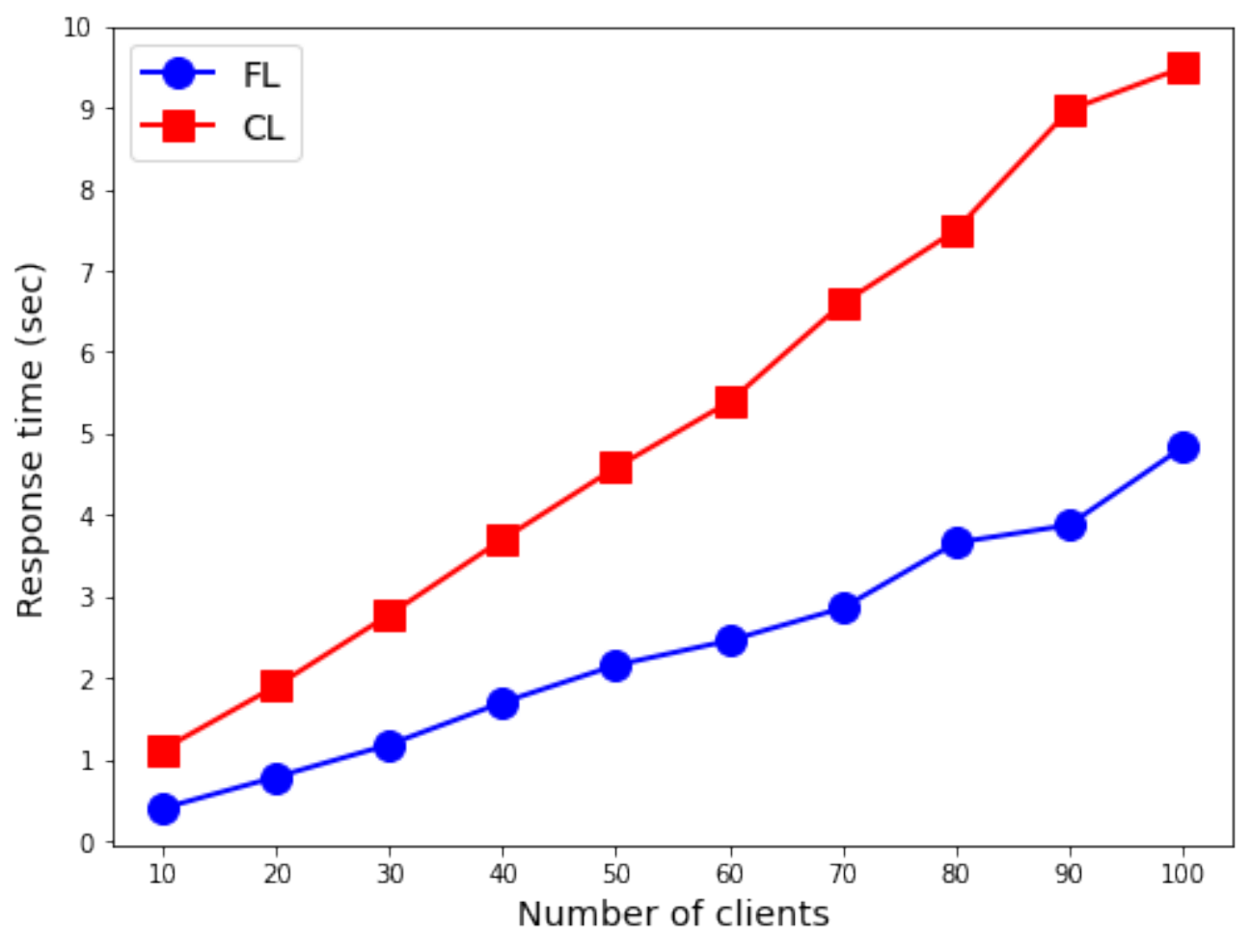}
    \caption{The response time under different clients: FL vs. CL.}
    \label{fig:cl_vs_lo_command_time}
\end{figure}

Figure \ref{fig:cl_vs_lo_command_time} compares how the response time varies with the number of clients in FL and CL.
We measure the response time from $T_C$ to $T_M$ for activities classified by a machine learning model.
The response time in FL is 0.4 seconds for 10 clients.
On the other hand, in CL, we observe that it took 1.1 seconds to complete the classification and IoT control job.
For 100 clients, it took 4.8 seconds with FL and 9.5 seconds with CL, resulting in the difference of 4.7 seconds.
The response time of CL under many clients is slow because the overhead of large file transmission and training increases to waste the computation resources of a CL server.

\begin{table}
  \caption{The response time for five user activities under different IoT devices (seconds).}
  \label{tab:responsetime_different_IoT_device}
  \centering
  \resizebox{0.48\textwidth}{!}{%
  \begin{tabular}{llcccccc}
    \toprule
    User activity & IoT device & $T_C$ & $T_F$ & $T_A$ & $T_M$ & $T_R$ & Total \\
    \midrule
    Reading & Philips Hue & 0.37 & - & 0.38 & - & 0.06 & 0.81 \\
    Drinking water & Google Nest Hub & 0.39 & - & 0.38 & - & 2.32 & 3.09 \\
    Using laptop & TP-Link & 0.39 & - & 0.38 & - & 0.8 & 1.57 \\
    Using mobile phone & TP-Link & 0.37 & - & 0.39 & - & 0.82 & 1.58 \\
    Washing dishes & Google Nest Hub & 0.39 & 0.64 & 0.38 & 0.01 & 12.63 & 14.05 \\
    \bottomrule
\end{tabular}}
\end{table}

Table \ref{tab:responsetime_different_IoT_device} shows the response time for five user activities.
In local control with FL, the response time is 0.81 seconds for the reading event; 3.09 seconds for the drinking water event; 1.58 seconds for events using laptop and mobile phone.
On the other hand, in remote control with CL, the response time for the washing dish event that plays YouTube on Google Nest Hub is 14.05 seconds.

\subsection{Accuracy and Privacy}
In this section, we perform two experiments.
First, we compare the accuracy of the FL models with and without transfer learning using TensorFlow Hub\footnote{https://www.tens0orflow.org/hub}.
Second, we investigate the impact of differential privacy on the accuracy and privacy of the FL model with TensorFlow Privacy\footnote{https://github.com/tensorflow/privacy}.

\subsubsection{Federated Learning with Transfer Learning vs. Federated Learning without Transfer Learning}
\begin{figure}
    \centering
    \includegraphics[width=0.45\textwidth]{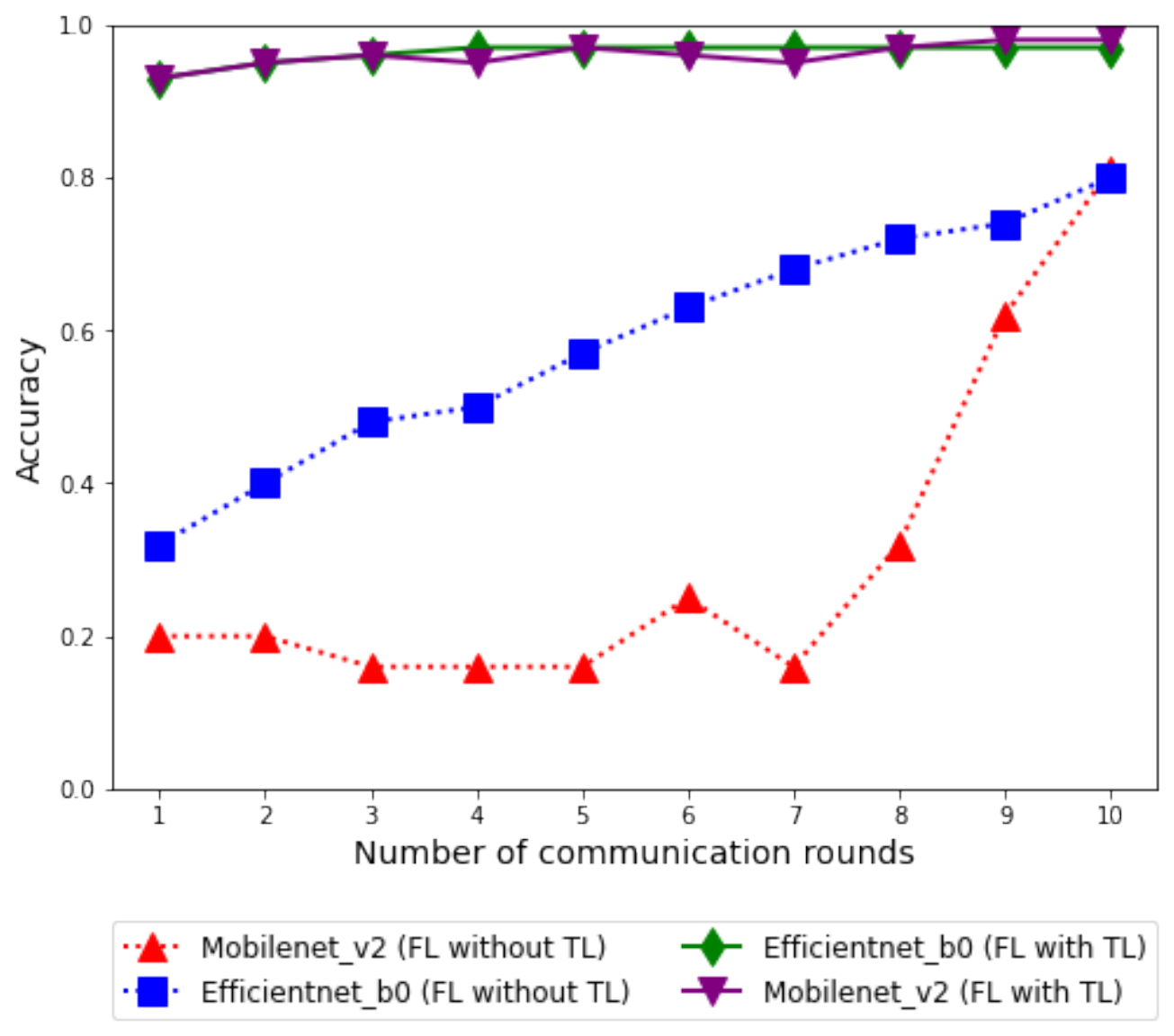}
    \caption{Accuracy of federated learning with or without transfer learning.}
    \label{fig:different_ftl_fntl}
\end{figure}

In Fig. \ref{fig:different_ftl_fntl}, we observed that federated learning with transfer learning (FL with TL) outperforms federated learning (FL without TL).
FL without TL starts with low accuracy due to insufficient data.
The accuracy of the FL with TL is 73\% higher than the FL without TL in the first round and 17\% higher in the 10$_{th}$ round.
Compared to the FL without TL, the FL with TL quickly achieves high performance.

\subsubsection{Federated Learning with Differential Privacy}

We train the model using a differentially-private stochastic gradient descent (DP-SGD) optimization algorithm.
As shown in Table \ref{tab:privacy_parameters}, we set the parameters to examine the performance of FL with TL.
$\epsilon$ is the privacy loss metric or privacy budget to measure the strength of privacy, and the probability of accidentally leaking information is $\delta$, which limits the probability that privacy is not guaranteed.
Moreover, we limit the exposure of personal information by setting the noise multiplier and the clipping threshold.

\begin{table}
  \caption{Sample parameters regarding privacy constraints. \texttt{num\_microbatches} is 32 (equal to \texttt{batch\_size}), and $\delta$ is $10^{-4}$ ($\delta < 1/n$, n is the number of training examples).}
  \label{tab:privacy_parameters}
  \centering
  \resizebox{0.38\textwidth}{!}{%
  \begin{tabular}{ccc}
    \toprule
    Noise multiplier & Clipping threshold & $\epsilon$ \\
    \midrule
    0.3 & 0.5 & 62.5 \\
    0.5 & 0.7 & 10.9 \\
    1.3 & 1.5 & 0.9 \\ 
    - & - & Non-private \\ 
    \bottomrule
\end{tabular}}
\end{table}

\begin{figure}
    \centering
    \includegraphics[width=0.45\textwidth]{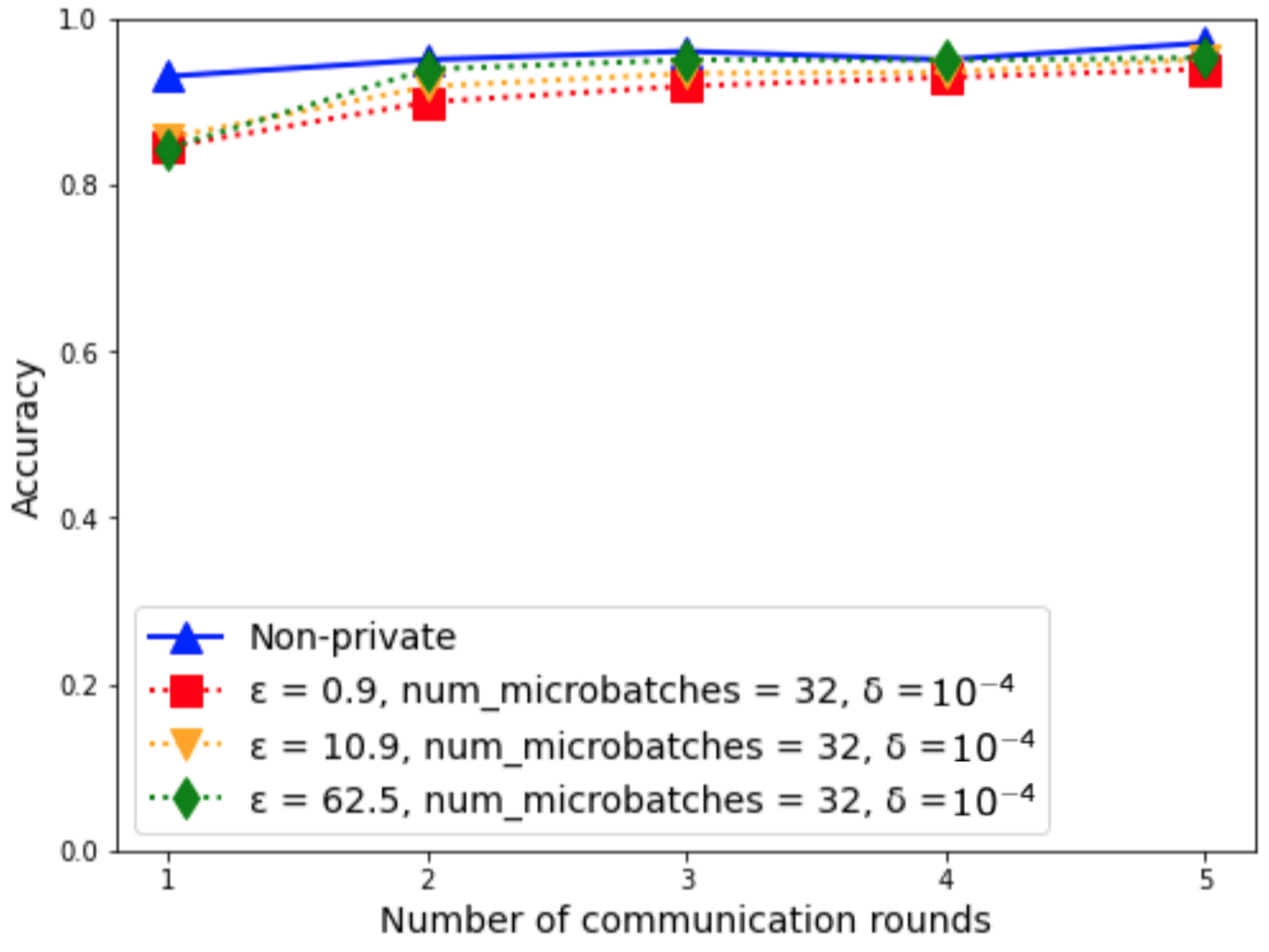}
    \caption{The accuracy with various privacy protection levels for FL with TL model (MobilenetV2).}
    \label{fig:ftl_dp}
\end{figure}

Figure \ref{fig:ftl_dp} shows the accuracy of the FL with TL model (MobilenetV2) with different levels of protection ($\epsilon$ = 0.9, $\epsilon$ = 10.9, and $\epsilon$ = 62.5).
In this experiment, we calculate the $\epsilon$ value from the parameters in Table \ref{tab:privacy_parameters}.
Since $\delta$ is set to be less than the inverse of the number of training data in privacy, we set $\delta$ to $10^{-4}$ in our experiment.
We can observe that as $\epsilon$ decreases, the level of privacy protection becomes high due to noise.
For MobilenetV2, the accuracy of the model with $\epsilon$ of 0.922 in the final round is 93\%, which is 2\% different from the model with 10.9.
In addition, we observe a slight difference of 4\% compared to the model without DP.
In the last round, the 95\% accuracy of the model with $\epsilon$ of 0.9 results in 3\% difference compared to the 98\% accuracy of the simple FL model without DP.
This means that our FL with TL and DP model can classify user activities even if we set the highest privacy strength ($\epsilon$ = 0.9) in our experiment.

\section{Conclusion and future work}
In this paper, we present a local control method for federated learning home IoT services that minimize the end-to-end response time.
The local control can minimize the end-to-end response time because there is no communication overhead with the cloud server.
In the learning process, the FL client directly trains the individually collected data and sends the results to the federated server.
We apply transfer learning to the federated learning model to improve the user context classification model accuracy due to insufficient data.
We also evaluate federated learning methods using differential privacy applied to provide improved privacy protection against model inversion attacks.

In future work, we plan to extend the IoT service of federated learning to various IoT devices and user activities.
We need a way to train models with scalable user activity for IoT devices.
This requires experimentation with real users' activities so that they can be generalized to federated learning environments.
We believe that a crowd-sourcing test that uploads an image of an activity by a user should also be developed as a method.
In addition, we consider the use of personal information in public places that value personal information, such as rest rooms and toilets, rather than in an environment where IoT devices are individually controlled.

\begin{acks}
This work was supported by Institute for Information \& communications Technology Planning \& Evaluation(IITP) grant funded by the Korea government (MSIT)(No.2019-0-01343, Training Key Talents in Industrial Convergence Security) and this work was supported by Institute of Information \& communications Technology Planning \& Evaluation(IITP) grant funded by the Korea government(MSIT) (No.2020-0-00901, Information tracking technology related with cyber crime activity including illegal virtual asset transactions).
Corresponding author is Youngseok Lee.
\end{acks}

\bibliographystyle{ACM-Reference-Format}
\bibliography{sample-bibliography} 

\end{document}